\documentclass[
]{ceurart}

\usepackage{listings}
\usepackage{graphicx}
\usepackage{hyperref}
\usepackage{subcaption}
\usepackage{placeins}

\graphicspath{{figures/}}

\usepackage{soul}
\usepackage{xcolor}

\usepackage{xspace}

\usepackage{adjustbox}
\usepackage{tabularx}
\usepackage{booktabs}
\usepackage{array}

\usepackage{float}

\definecolor{asteroid1}{HTML}{0A9F9D}
\definecolor{asteroid2}{HTML}{CEB175}
\definecolor{asteroid3}{HTML}{E54E21}
\definecolor{asteroid4}{HTML}{6C8645}
\definecolor{asteroid5}{HTML}{C18748}

\definecolor{asteroid6}{HTML}{C52E19}
\definecolor{asteroid7}{HTML}{AC9765}
\definecolor{asteroid8}{HTML}{54D8B1}
\definecolor{asteroid9}{HTML}{b67c3b}
\definecolor{asteroid10}{HTML}{175149}
\definecolor{asteroid11}{HTML}{AF4E24}

\definecolor{asteroid12}{HTML}{FBA72A}
\definecolor{asteroid13}{HTML}{D3D4D8}
\definecolor{asteroid14}{HTML}{CB7A5C}
\definecolor{asteroid15}{HTML}{5785C1}

\newcommand{\seqdream}{\setulcolor{asteroid12}\ul{\textit{Sequence Dreaming}}\xspace}
\newcommand{\seqdreamshort}{\setulcolor{asteroid12}\ul{\textit{SD}}\xspace}
\newcommand{\seqdreamwithout}{\textit{Sequence Dreaming}\xspace}

\newcommand{\underlined}[2]{\setulcolor{#1}\ul{#2}\xspace}

\begin{document}

\copyrightyear{2024}
\copyrightclause{Copyright for this paper by its authors.
  Use permitted under Creative Commons License Attribution 4.0
  International (CC BY 4.0).}

\conference{TempXAI@ECML-PKDD'24: Explainable AI for Time Series and Data Streams Tutorial-Workshop, Sep. 9\textsuperscript{th}, 2024, Vilnius, Lithunia}

\title{Finding the DeepDream for Time Series:\\Activation Maximization for Univariate Time Series}

%
\author[1]{Udo Schlegel}[%
orcid=0000-0002-8266-0162,
email=u.schlegel@uni.kn,
url=https://www.vis.uni-konstanz.de/mitglieder/schlegel/,
]
\cormark[1]
\address[1]{University of Konstanz,
  Universitätsstraße 10, Konstanz, 78464, Germany}

\author[1]{Daniel A. Keim}[%
orcid=0000-0001-7966-9740,
email=daniel.keim@uni.kn,
]

\author[1]{Tobias Sutter}[%
orcid=0000-0003-1226-6845,
email=tobias.sutter@uni.kn,
]

\cortext[1]{Corresponding author.}

\begin{abstract}
Understanding how models process and interpret time series data remains a significant challenge in deep learning to enable applicability in safety-critical areas such as healthcare. 
In this paper, we introduce \seqdream, a technique that adapts Activation Maximization to analyze sequential information, aiming to enhance the interpretability of neural networks operating on univariate time series. 
By leveraging this method, we visualize the temporal dynamics and patterns most influential in model decision-making processes. 
To counteract the generation of unrealistic or excessively noisy sequences, we enhance \seqdream with a range of regularization techniques, including exponential smoothing. 
This approach ensures the production of sequences that more accurately reflect the critical features identified by the neural network.
Our approach is tested on a time series classification dataset encompassing applications in predictive maintenance.
The results show that our proposed \seqdream approach demonstrates targeted activation maximization for different use cases so that either centered class or border activation maximization can be generated.
The results underscore the versatility of \seqdream in uncovering salient temporal features learned by neural networks, thereby advancing model transparency and trustworthiness in decision-critical domains. 
\end{abstract}


\maketitle

\section{Introduction}


Techniques such as Activation Maximization~\cite{simonyan_deep_2014} and DeepDream~\cite{mordvintsev_deepdream_2015} have emerged as approaches to make the complex inner workings of deep neural networks (DNNs) more transparent. 
These methods illuminate the black box of neural networks by visualizing what neural networks learn, significantly improve model interpretability, and provide valuable insights into diagnosing model behavior~\cite{olah_feature_2017}. 
Activation Maximization focuses on identifying the input patterns that maximize the response of specific neurons or layers, revealing the features and patterns a model perceives as most salient~\cite{simonyan_deep_2014}. 
Meanwhile, DeepDream leverages the layers of neural networks to generate intricate, dream-like (unreal-looking) images that highlight the learned features in a visually compelling way~\cite{mordvintsev_deepdream_2015}. 
Together, these techniques enhance our understanding of how neural networks process information and guide the development of more transparent, effective, and interpretable AI systems. 
Through the lens of Activation Maximization and DeepDream, we can unravel the intricacies of neural networks, paving the way for advancements in AI that are comprehensible~\cite{olah_feature_2017}.


In the evolving landscape of neural network interpretation, adapting Activation Maximization for time series data extends the understanding of how deep learning models perceive and process temporal information. 
\seqdream enables visualization of the intricate temporal features and dynamics the network has learned to recognize by manipulating time series data to amplify the patterns that maximally or targeted activate specific neurons within a model. 
This method sheds light on the model's decision-making process and unveils the temporal sequences and patterns deemed most significant by the neural network. 
In doing so, \seqdream bridges the gap between the opaque decision-making of deep learning models and the tangible insights they derive from sequential data, offering another lens through which to interpret and refine models trained on time series similar to shapelet learning~\cite{theissler_explainable_2022}. 
This approach promises to enhance model transparency, facilitate diagnostic analysis, and inspire the development of models for handling the complexities of sequential data analysis.


\begin{figure}[ht]
    \centering
    \begin{subfigure}[b]{\textwidth}
        \centering
        \includegraphics[trim=0.7cm 0.2cm 0.7cm 0.2cm,clip,width=\textwidth]{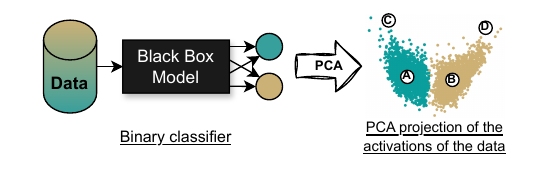}
        \caption{A black box model gets trained on binary data to classify between two classes. The logits before the softmax can then be projected again with PCA to generate a visual representation of the predictions. The classifier splits the predictions into two clusters with the logits. However, we are interested in the different parts of the clusters. What are the cluster centers (A) or (B)? How do the cluster edges look like (C) or (D)? How can we recreate the data of these interesting regions?}
        \label{fig:general_motivation}
    \end{subfigure}
    
    \vspace{1em}
    
    \begin{subfigure}[b]{\textwidth}
        \centering
        \includegraphics[trim=0.7cm 0.2cm 0.7cm 0.2cm,clip,width=\textwidth]{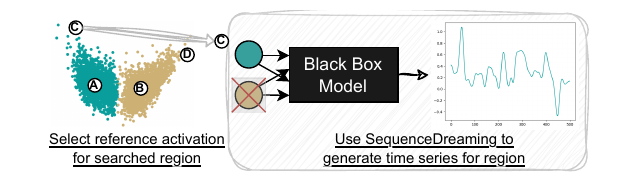}
        \caption{Select a reference activation such as a class cluster center (A) or a maximization such as (C). Here, we want to create a maximization, so (C) is selected. We also want the selected activation of the class. Thus, the other class activation gets deactivated. Through gradient optimization, the input of the black box model gets slowly adapted to generate a time series for the selected region.}
        \label{fig:general_approach}
    \end{subfigure}
    \caption{General approach split up into two pipelines. First, a projection of the activations of the data into 2D to find interesting regions to focus on. Second, the \seqdream approach generates time series in selected regions of interest to find salient features of the model.}
    \label{fig:general}
\end{figure}

In this paper, we introduce \seqdream, an adaptation of the activation maximization techniques for DeepDream~\cite{mordvintsev_deepdream_2015,mahendran_understanding_2015} explicitly tailored for time series data. 
By applying this method, we aim to enhance the interpretability of deep learning models that process time series data, shedding light on the temporal dynamics or patterns these models capture. 
To refine and control the generation of artificial time series that maximize neuron activations, we extend \seqdream with a suite of regularization techniques, including an \(\alpha\)-norm, total variation, time point smoothness, Gaussian smoothing, and random noise reinitiation to produce more realistic and informative visualizations. 
These modifications are designed to mitigate overfitting to noise and emphasize the underlying patterns critical for model decisions. 
We test our approach on a time series classification dataset tackling predictive maintenance.
Our methodology advances the field of model interpretability for time series analysis and offers a framework for improving the transparency and trustworthiness of models deployed in critical decision-making domains.

Source code for \seqdream and results of the experiments are online available at:
\makebox[\textwidth]{\href{https://github.com/visual-xai-for-time-series/sequence-dreaming}{https://github.com/visual-xai-for-time-series/sequence-dreaming}}

\section{Related Work}


In explainable artificial intelligence (XAI), elucidating the operational intricacies of machine learning algorithms, particularly for time series data, is imperative for advancing the field's theoretical and practical applications. 
Theissler et al.~\cite{theissler_explainable_2022} delineate a comprehensive framework for examining time series XAI, emphasizing the necessity of addressing interpretability at multiple analytical levels. 
This multi-tiered perspective is instrumental in unraveling the complex dynamics and temporal dependencies inherent in time series data, thereby facilitating a deeper understanding of model behavior.

Activation Maximization, within this context, emerges as a crucial methodology for probing the internal representations developed by neural networks during the training process~\cite{simonyan_deep_2014}. 
It accomplishes this by optimizing input signals to obtain maximal responses from specific neurons or layers, effectively revealing the features and patterns deemed most significant by the model for its decision-making processes. 
DeepDream enhances Activation Maximization by incorporating regularization techniques into the optimization process, guiding the model to modify inputs (typically images) to emphasize the learned features while maintaining or even enhancing the visual coherence and interpretability of the outputs~\cite{mordvintsev_deepdream_2015}. 


Thus, DeepDream not only highlights the features that activate certain neurons but does so in a manner that produces aesthetically intriguing and richly detailed images, thereby making the abstract concepts learned by the network more tangible and understandable to humans~\cite{mahendran_understanding_2015}.
One of the earliest examples of reconstructing images based on a neuron activation is provided by Yosinski et al.~\cite{yosinski_understanding_2015} incorporating Gaussian blur, cropping by pixel contribution, and cropping by pixel norm without including a loss function.
Nguyen et al.~\cite{nguyen_deep_2015} demonstrate that DNNs can be induced to make high-confidence predictions for images that are either nonsensical or unrecognizable to human observers, highlighting significant vulnerabilities in the models' interpretability.

Integrating Activation Maximization techniques within time series analysis significantly enhances our understanding of neural network predictions, thereby improving the transparency and accountability of these models for critical applications, for instance, in Schlegel et al.~\cite{schlegel_visual_2023}. 
However, as highlighted in the literature, applying Activation Maximization to time series data presents unique challenges, not least due to the complexity of temporal data~\cite{yoshimura_preliminary_2019}. 
Various studies, including those by Yoshimura et al.~\cite{yoshimura_preliminary_2019} and ~\cite{yoshimura_understanding_2021}, have explored alternative approaches, such as utilizing the spectral domain and Fourier transformations on short time series, often focusing beyond the last layer of the network. 
Meanwhile, Ellis et al.~\cite{ellis_novel_2021} have proposed a novel method that involves perturbing the spectral domain to achieve Activation Maximization, although these results are often overly focused on periodicity. 
This indicates a pressing need for a more systematic approach to effectively apply Activation Maximization in the context of time series data, aligning with the broader objectives of explainable and interpretable machine learning models as advocated by the XAI community~\cite{yoshimura_preliminary_2019}.


\section{Activation Maximization}


Activation Maximization emerges as a promising technique designed to reveal the features most salient to individual neurons within a trained neural network, enabling interpretability~\cite{mordvintsev_deepdream_2015}. 
Post-training, the focus shifts to decoding the learned representations by identifying the inputs that provoke maximal activation in specific neurons. 
This process, integral to Activation Maximization, aims to uncover the features or patterns to which a neuron is most responsive by iteratively refining the input to maximize a neuron's activation, providing deeper insights into the model's internal representations~\cite{simonyan_deep_2014}.
However, applying this technique to time series data introduces novel challenges necessitating regularization strategies.

In time series classification, we can define a \textit{time series} by \( ts = (t_1, t_2, \dots, t_m) \in \mathbb{R}^{m\times d} \) as an ordered set of \( m \) real-valued observations (or time steps), with dimensionality \( d \)~\cite{theissler_explainable_2022}.
For \textit{univariate} time series, we have \( d = 1 \) and thus our \( ts \in \mathbb{R}^{m} \).
Given a trained model \( M \) and a target class \( c \), the activation maximization is described as follows: 
Consider a score function $S_c:\mathbb{R}^m\to\mathbb{R}$ and let \( S_c(I) \) denote the score of the class \( c \) from \( M \) computed by the classification layer of the model for an input \( I = (x_1, x_2, \dots, x_m) \in\mathbb{R}^m \), e.g., \( I = ts \) and thus \( S_c(ts) \).
Thus, our previous time series \( ts \) works as an input for the model \( M \).
Next, we want to find an \( L_2 \)-regularised input, such that the score \( S_c \) is maximized, i.e.,
\begin{equation}
\max_{ts} S_c(ts) - \lambda \|ts\|^2_2,
\label{eq:argmax-loss}
\end{equation}
where \( \lambda > 0 \) is a regularisation parameter. 
By employing the back-propagation technique, we can identify an input, referred to as \( ts \), that is locally optimal in terms of the model's criteria.
Rather than modifying the network's weights, we hold them constant at the values established during the training phase and instead focus our optimization on the input \( ts \) itself.
In our case, we focus on \( S_c(ts) \) and, in the first step, remove the \( L_2 \)-regularised input, focusing directly on the maximization of the activation score of the class, i.e., \( \lambda = 0 \).
For this approach, we use a gradient ascent to fine-tune the input to increase the activation.

\section{Regularization Tricks}
Regularization plays a crucial role in Activation Maximization, primarily to ensure the generation of interpretable, meaningful, and visually coherent inputs~\cite{yosinski_understanding_2015}. 
It prevents the optimization process from overfitting to noise, which would otherwise lead to unrecognizable patterns that maximize neuron activation but lack relevance. 
By introducing regularization, such as Gaussian blur~\cite{yosinski_understanding_2015}, the process is guided towards producing aesthetically pleasing inputs closer to the distribution of natural inputs, enhancing the interpretability of the results. 
This ensures that the generated inputs reflect genuine features the network learns from real-world data, offering clearer insights into its internal representations and improving our understanding of its decision-making processes.

\subsection{Without regularization on the loss}
Exploring regularization techniques that do not alter the loss function presents an approach in optimization we first want to explore, particularly within Activation Maximization using gradient ascent. 
Thus, we collect and reformulate the approaches from the literature~\cite{yosinski_understanding_2015,yoshimura_preliminary_2019,ellis_novel_2021} to time series. We focus on the following approaches:


\textbf{Clamping to borders}, as described by Yosinski et al.~\cite{yosinski_understanding_2015}, ensures that the optimized input does not venture beyond the predefined input space, maintaining realism and coherence. 
This technique effectively keeps the activation maximization process within the bounds of the training data distributions.

\textbf{\( L_2 \) decay}, another technique highlighted by Yosinski et al.~\cite{yosinski_understanding_2015}, imposes a regularization term that penalizes high-frequency noise in the generated input, promoting smoother and more interpretable features. 
This regularization helps focus on the essential features that the neuron detects rather than artifacts.

\textbf{Random scaling} introduces variability in the time points of the input during the optimization process, encouraging the network to identify and amplify scale-invariant time points. 
This technique enriches the diversity of patterns that activate the neurons, showcasing the model's robustness to scale variations.

\textbf{Moving average smoothing} is applied to the input to mitigate rapid fluctuations, ensuring that the input generation progresses smoothly toward enhancing meaningful patterns. 
This approach helps stabilize the time points, making it less susceptible to local optima.

\textbf{Exponential smoothing}, when applied to the input during the optimization process, emphasizes the significance of smoother inputs by assigning them greater weights. 
This method adapts the input dynamically, ensuring that the activation maximization is finely tuned based on the most recent trends and patterns in the data, thereby fostering a more responsive and effective approach to highlighting the neuron's preferences.

\textbf{Gaussian blur filter}, as employed by Yosinski et al.~\cite{yosinski_understanding_2015}, aids in reducing high-frequency noise across the optimization iterations, thereby focusing the model's attention on broader, more significant patterns. 
This technique contributes to the production of more visually appealing and interpretable inputs.

\textbf{Random reinitiation} of the input, if the optimization process shows no significant change, acts as a reset mechanism to escape from potential plateaus in the activation landscape. 
This strategy prevents stagnation, ensuring continuous exploration for more effective stimuli that maximally activate the target neuron.

\textbf{\underlined{asteroid15}{Intuitive Explanation} -- }%
Most of the approaches mentioned above are straightforward in their methodology. 
Generally, they aim to regularize the creation of the time series without altering the loss function, focusing solely on increasing the activation of the selected neuron. 
This often necessitates significant smoothing since there are no constraints on the loss function and, consequently, the gradients.

\subsection{With regularization on the loss}

After we adopt a regularization approach without modifying the loss function, utilizing gradient ascent as demonstrated by Yosinski et al.~\cite{yosinski_understanding_2015}, we transition the approach imposing regularization directly on the loss function, thereby altering our optimization strategy.
We no longer perform a full activation maximization but aim to approximate a specific target activation in advance.
Thus, we shift from gradient ascent to gradient descent, employing traditional optimizers and thus integrating a specific loss term with, for example, stochastic gradient descent with momentum to refine our optimization process further.


In extending~\autoref{eq:argmax-loss}, we incorporate total variation regularization, as outlined by Simonyan et al.~\cite{simonyan_deep_2014}, to enhance the visual clarity and reduce noise in the generated input, which also enables smoothing for the input time series. 
To optimize this extended formulation, we employ the Adam optimizer~\cite{kingma_adam_2014}, known for its efficiency in handling sparse gradients and adaptive learning rates, complemented by weight decay for improved regularization on the input, following the approach suggested by Mahendran and Vedaldi~\cite{mahendran_understanding_2015}. 
Additionally, we meticulously adjust the normalization parameters for the \( L_2 \)-Regularization (weight decay on the input) and Total Variation (on the input) and fine-tune the weighting of the loss terms to identify the most effective settings for our optimization objectives, ensuring a balanced and nuanced approach to maximizing activation as closely to our target as possible.

Thus, we get the formula from Mahendran and Vedaldi~\cite{mahendran_understanding_2015}
\begin{equation}
\min_{ts} \frac{|S_c(ts) - S_c(T)|^2}{|S_c(T)|^2} + \lambda_{\alpha}\cdot \|ts\|^{\alpha}_{\alpha} + \lambda_{\beta}\cdot TV(ts,\beta),
\label{eq:argmax-losswith}
\end{equation}
with \( S_c(ts) \) as the score of the class \( c \) for an input \( ts \) and \( S_c(T) \) as the score of the class \( c \) for a target \( T \). 
The target \( T \) and \( S_c(T) \) as a reference for the activation we want to achieve, as mentioned before.
The parameter \( \alpha \) changes the input range to be encouraged to stay within an interval if set to large values $> 2$, commonly set to  \( \alpha = 6 \)~\cite{mahendran_understanding_2015}.
The parameter \( \beta \) controls the total variation factor and, thus, how similar the inputs in the neighborhood should be.
The parameter is normally \( \beta > 1 \) so that the weighting of the error between the time points can be adjusted.
In our case, we want a confident prediction of a sample toward the class \( c \) after the softmax, e.g., \( M(T) = [0, 1] \) for two classes and \( c = 2\) for \( c\in \{1, 2\}\).

\( TV \) corresponds to the total variation adapted and approximated with
\begin{equation}
TV(ts,\beta) = \sum_{i=1}^{m} (t_{i+1} - t_{i})^{\beta}
\label{eq:argmax-tv}
\end{equation}
as seen in Mahendran and Vedaldi~\cite{mahendran_understanding_2015} adapted to time series with just one dimension.
The parameter \( \lambda_{\alpha} \)  weight the \( L_2 \) regularization term, while \( \lambda_{\beta} \) weights the \( TV \).

\textbf{\underlined{asteroid15}{Intuitive Explanation} -- }%
The loss terms can be intuitively explained as follows: 
The first part ensures that the input score gradually moves toward the desired target score, where the activation of the input should resemble the activation of the target. 
In our case, the target is a high activation above two or three times above the average of the activations of the train data.
The second part keeps the input values within a set range, provided that the parameter \( \alpha \) is chosen carefully. 
Here, the normalization focuses on making outlier values more averaging out with the other values.
Lastly, the final part ensures that the neighborhood has similar values without outliers, as the $TV$ smooths the data with a \( \beta \) value greater than or equal to 1.
With a high \( \beta \), errors between time points are heavier weighted, thus minimizing the equation forces to lower the difference between time points.


\subsection{\seqdreamwithout combining both}


\seqdream combines the most promising regularization techniques from previous work on images with specialized ones on time series to generate an approach for time series deep learning models. 
By extending~\autoref{eq:argmax-losswith} with an additional smoothness factor similar to TV but scaled for time series lengths, \seqdream achieves a more refined regularization process. 
This method employs a combination of the new loss extension, Gaussian blur, clamping, and random noise to ensure effective regularization, particularly when the loss is not significantly changing. 
Furthermore, \seqdream transitions from using Adam to a gradient descent method without momentum or other improvements, necessitating a few more optimization steps to achieve the desired results.


First, we extend~\autoref{eq:argmax-losswith} by adding an additional regularization term
\begin{equation}
\min_I \frac{|S_c(ts) - S_c(T)|^2}{|S_c(T)|^2} + \lambda_{\alpha}\cdot \|ts\|^{\alpha}_{\alpha} + \lambda_{\beta}\cdot TV(ts,\beta) + \lambda_{sm}\cdot SM(ts),
\label{eq:sequence-dreaming-eq}
\end{equation}
where ($SM$)oothness \( SM(ts) \) is defined as
\begin{equation}
SM(ts) = \frac{1}{m-1} \cdot \sum_{i=1}^{m-1} |t_{i+1} - t_{i}|.
\label{eq:sq-smooth}
\end{equation}
$SM$ takes the input, calculates the discrete difference between time points, uses the absolute value of the difference, and sums these up.
The normalization of the length enables an equal weighting between every time point.

\textbf{\underlined{asteroid13}{Why $SM$?} -- }
The inclusion of two smoothing factors, despite $TV$'s inherent smoothing capabilities, arises from their distinct focuses and benefits. 
$TV$ primarily aims to minimize errors between individual time points, emphasizing the coherence of each point in the time series. 
In contrast, the second smoothing factor, $SM$, targets the entire time series, enhancing overall normalization and providing a broader perspective on data regularization. 
While $TV$ hones in on reducing local discrepancies, $SM$ ensures a holistic approach, balancing the series as a whole and easing the optimization path in the loss landscape. 
This dual-faceted approach enables a more comprehensive smoothing process, addressing micro-level and macro-level anomalies within the time series.

Another important discussion point is the starting point of the optimization. 
While it is common to begin with completely black or white images, some approaches use the mean of a chosen dataset or random values~\cite{mahendran_understanding_2015}. 
We provide a more directed approach to the starting input for optimization as time series starting at zero are often standardized and already incorporate biases. 
Specifically, we utilize the activations seen in~\autoref{fig:general_approach} to guide the search for the initial input. 
Different regions of the activations in~\autoref{fig:general_approach} can be used as starting points to steer optimization based on desired outcomes. 
For instance, in most cases, such as (A) in~\autoref{fig:general_approach}, starting with the closest sample to the mean activation of the dataset proves beneficial. 
This method enables the process to commence with a solid activation baseline, which can then be further optimized. 
Thus, \seqdream adapts an existing sample to align with the model's preferences, thereby maximizing the activation of the selected neuron.

To search for the most fitting hyperparameters, we perform a grid search due to the extensive range of possible hyperparameters that require fine-tuning, such as \( \lambda_{sm} \). 
In some instances, we have an idea of the optimal direction for hyperparameters, such as 
\( \alpha \). 
However, determining these parameters in many cases is not straightforward and is highly sensitive to the dataset, particularly the weighting parameters for the loss term \( \lambda_{sm} \). 
Through empirical testing, we discovered that while some parameters are sensitive, we can generally provide certain guidance towards the hyperparameters discussed in the evaluation section.

\textbf{\underlined{asteroid15}{Intuitive Explanation} -- }%
\seqdream enhances the previous loss function with an additional smoothing factor to ensure that every time point throughout the time series has similar importance. 
By focusing further on smoothing, we aim to optimize the resulting time series into a smooth version without large outliers, making it more realistic. 
Ultimately, this optimization results in a smooth time series with only patterns and outliers at important time points. 
Our implementation techniques further steer the direction to closely match the target activation by adding noise if the activation becomes too large and risks overshooting.

\section{Evaluation of Results and Discussion}

Evaluation is crucial because different approaches can generate activation maximization samples that are not always plausible for our dataset. 
Outside the projected data, these samples may still maximize neuron activation but in a different region, as illustrated in~\autoref{fig:general_approach}. 
Ellis et al.~\cite{ellis_novel_2021} assess their activation maximization approach by comparing the activation distribution against the data, performing a visual evaluation, analyzing the frequency domain, and examining frequency importance. 
Similarly, we will use a visual evaluation to present the generated time series. 
Subsequently, we will employ measures for outlier detection to assess the plausibility of the generated activation maximization input toward the model and the data. 
Finally, we will integrate out-of-distribution analysis and visual evaluation with distribution plots to compare \seqdream with other approaches.


\begin{adjustbox}{center,caption={Comparison of the generated activation maximization time series with their corresponding predictions and activations with the two classes (\underlined{asteroid1}{first class}, \underlined{asteroid2}{second class}). Ellis et al.~\cite{ellis_novel_2021} demonstrate plausible time series. However, our \seqdream (SD) comes up with totally different ones while smoothing the without-loss and with-loss results.},label={fig:comparison-line-plots},nofloat=figure,vspace=\bigskipamount}
\includegraphics[width=0.95\textwidth]{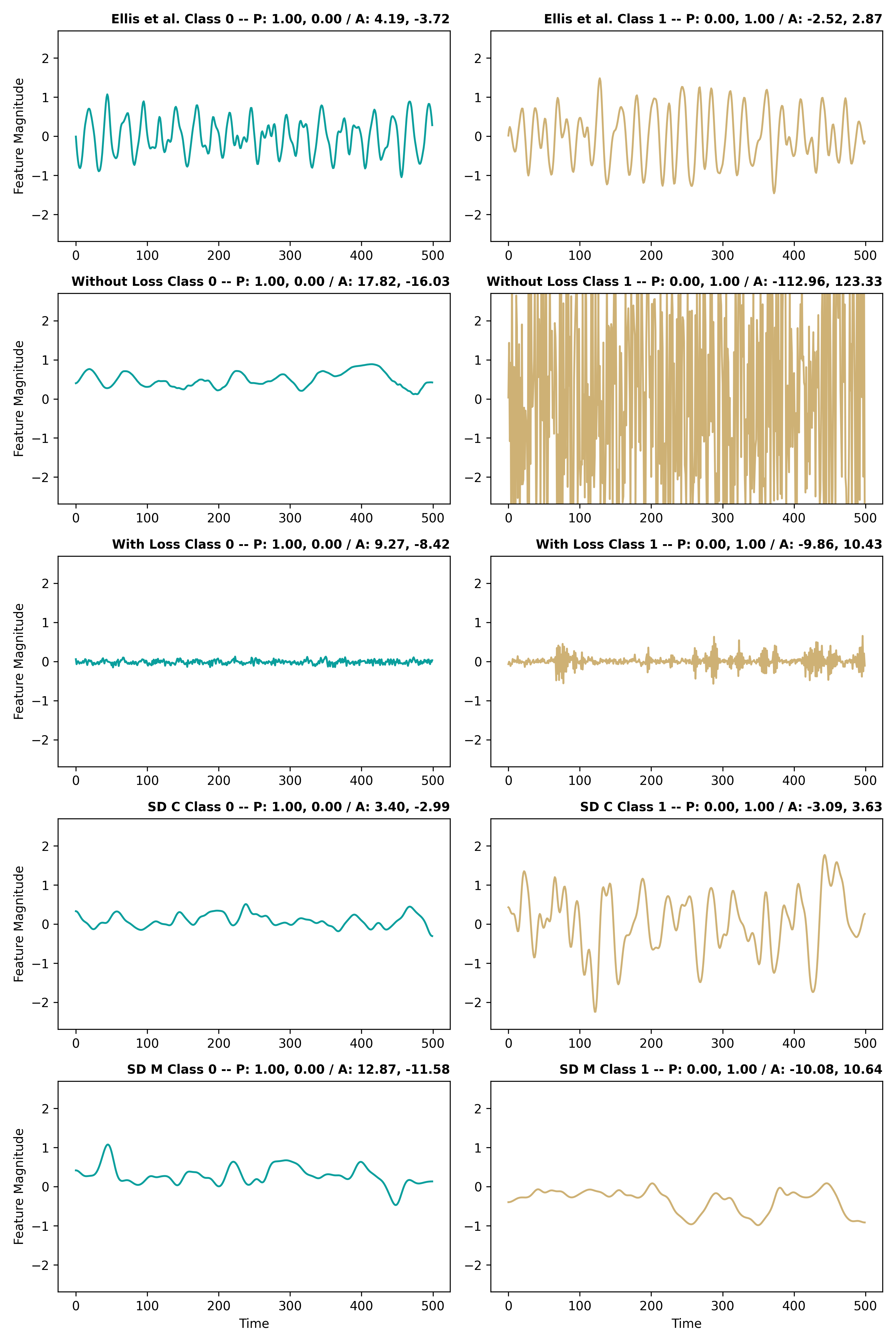}
\end{adjustbox}


The evaluation is conducted on the \textit{FordA} dataset, a well-established univariate time series classification benchmark for anomaly detection in car engines. 
This dataset consists of time series samples with 500 time points each, comprising 3,601 samples for training and 1,320 for testing. 
However, all datasets available from the UCR benchmark repository~\cite{dau_ucr_2019} are compatible with the \seqdream source code available online.
The model employed is a conventional \textit{ResNet} architecture featuring three ResNet blocks, each containing three Conv1D layers, followed by a final linear classifier layer. 
The model is trained using the Adam optimizer~\cite{kingma_adam_2014} over 500 epochs.
The model achieves an accuracy of \textit{0.99} on the training set and \textit{0.95} on the test set, approaching near \textit{state-of-the-art} performance.
For the hyperparameter search in \seqdream, a grid search is utilized with minimum and maximum values as outlined in~\autoref{tab:parameter}. 
While the current parameter ranges are already relatively narrow, they can be further reduced to accelerate the search process.
The hyperparameter ranges were determined based on literature references and empirical testing with respect to the loss function.



\begin{table}[htb]
\centering
\caption{Hyperparameter Search Space for \seqdream on the FordA Dataset using the ResNet Model. A grid search is applied, and the search space can be further reduced to streamline the search process and enhance the speed of the overall identification.}
\label{tab:parameter}
\begin{tabularx}{\textwidth}{XXXXXXXXX} 
\toprule
    & steps & lr & $\alpha$ & $\beta$ & $\sigma$ & $\lambda_\alpha$ & $\lambda_\beta$ & $\lambda_{sm}$ \\
\midrule
Min &   5    &  1e-2  &    4   &   1   &    3   &     1e-5     &     1e-5     &      1e-1     \\
Max &   100    &  1e1  &    6   &   2   &    6   &     1e-1     &     1e-1     &      5e-1    \\
\bottomrule
\end{tabularx}
\end{table}

\subsection{Visual Evaluation}

\autoref{fig:comparison-line-plots} visualizes the different approaches as line plots for direct comparison. 
At the top, the method name is displayed, followed by the prediction using the generated time series, and finally, the activation of the neurons in the selected layer. 
The left side always represents the first class, while the right side represents the second class.
The figure includes methods from Ellis et al.~\cite{ellis_novel_2021}, the mentioned approach without loss change, the approach with loss change, \seqdream on the class center, and the maximization using \seqdream.

Comparing the different approaches at a glance provides initial insights seen in~\autoref{fig:comparison-line-plots}. 
Ellis et al.~\cite{ellis_novel_2021} produce quite an appealing time series, but the use of FFT introduces some periodicity in the generated time series. 
The generated time series without a change in the loss results in high values for the \underlined{asteroid2}{second class (1)} and quite low values for the \underlined{asteroid1}{first class (0)}, indicating that this approach is not very effective. 
The time series generated with loss adjustment are not as smooth and tend to be somewhat noisy, particularly towards the \underlined{asteroid2}{second class}. 
In contrast, \seqdream generates convincing time series for the class center and maximization, with distinct differences between the classes. 
Notably, the \underlined{asteroid1}{first class} looks quite similar when \seqdream is targeted on the class center and class maximization, while the \underlined{asteroid2}{second class} results in noticeably different generated time series.

\subsection{Out-of-distribution evaluation}

After visually comparing the activation maximization results, we aim to assess their performance using various quantitative methods. 
The distributions of the class' activations for the training data are particularly interesting in this context to compare them to the generated time series. 
Different methods exist for evaluating data based on distributions, though we focus on outlier analysis to assess the quality of the generated time series towards the training data.

In our case, we employ outlier analysis using the Mahalanobis distance~\cite{mahalanobis_generalized_1936} on time series data and activations. 
The Mahalanobis distance measures the distance between a point and a distribution, considering the correlations between variables. 
It is calculated by determining how many standard deviations away a point is from the mean of the distribution, considering the covariance matrix to account for the data's spread and orientation. 
This makes it a useful approach for identifying outliers, as points that are far from the mean in terms of the Mahalanobis distance are likely to be anomalies.
We believe that applying this measure directly to the time series data is not ideal due to its diversity; only existing time series or those generated in the frequency domain would fit within the distribution. 
However, we do think this approach can effectively demonstrate how well the generated time series fit into the activations, revealing whether they are more on the border with high activation or more centered.

\begin{table}[htb]
\centering
\caption{Mahalanobis distance for the activations of the generated data and the training data activations. Min and Max correspond to the minimum and maximum distances of samples in the data based on the Mahalanobis distance for the data. Ellis et al.~\cite{ellis_novel_2021} perform quite well. However, \seqdream (SD) for the center works quite well, and also for the maximization, which needs to be more different than the previous ones to show that the neuron gets maximally activated.}
\label{tab:moa_act}
\begin{tabularx}{\textwidth}{ccccXXXX}
\toprule
Class &  Min &  Max &  Ellis et al.~\cite{ellis_novel_2021} &  Without Loss &  With Loss &  \seqdreamshort Center &  \seqdreamshort Max \\
\midrule
 0 &           0.11 &           5.18 &          1.36 &         13.86 &       3.92 &                 1.63 &                 8.41 \\
 1 &           0.10 &           4.61 &          1.38 &        137.74 &       3.54 &                 1.32 &                 3.53 \\
\bottomrule
\end{tabularx}
\end{table}

\textbf{Evaluation against activations from the training dataset -- }
\autoref{tab:moa_act} shows the Mahalanobis distance toward the activations of the training data compared to the generated time series by the different approaches. 
Min and Max correspond as a reference to the minimum and maximum distances of the samples in the training data itself.
Ellis et al.~\cite{ellis_novel_2021} performs quite well, positioning very nicely between the min and max. 
However, due to the small values, these are rather center class prototypes and not activation maximization. 
The approach without loss changes performs rather poorly.
In contrast, the approach with loss changes falls nicely between the min and max, even leaning more towards the borders. 
This indicates that the generated time series yields favorable results according to this measure. 
However, the visual evaluation shows that the results are not impressive.
\seqdream center presents values similar to Ellis et al.~\cite{ellis_novel_2021}, aligning well with the distribution but not reaching the borders. 
This outcome was expected since we generated these time series to be more centered, indicating that the approach works effectively based on this measure. 
Additionally, \seqdream max works quite well by providing larger numbers for the distance, suggesting that the generated time series are anomalies and somewhat borderline activations as intended.
Overall, our proposed \seqdream works very well for the activations seen in~\autoref{tab:moa_act}.


\begin{table}[hbt]
\centering
\caption{Mahalanobis distance for the time series training data and the generated time series. Min and Max correspond to the minimum and maximum distances of samples in the data based on the Mahalanobis distance for the data. Even Ellis et al.~\cite{ellis_novel_2021} with using the frequency domain for the generation can only generate class one time series in the time series training data distribution, which is very surprising. \seqdream (SD) cannot produce a time series in the distribution, which suggests that the model learns more abstract patterns for the classification. }
\label{tab:moa_ts}
\begin{tabularx}{\textwidth}{ccccXXXX}
\toprule
Class & Min &  Max &  Ellis et al.~\cite{ellis_novel_2021} &  Without Loss &   With Loss &  \seqdreamshort Center &  \seqdreamshort Max \\
\midrule
 0 &           1.23 &          55.53 &      23842.49 &  652852860.28 & 18435054.75 &         123537050.76 &         376120973.62 \\
 1 &           1.16 &          33.88 &         54.34 &   74930207.64 &  7966354.20 &          56484405.81 &         461642519.01 \\
\bottomrule
\end{tabularx}
\end{table}

\textbf{Evaluation against the training dataset -- }
\autoref{tab:moa_ts} presents the Mahalanobis distance from the original time series training data to the generated time series. 
Min and Max again correspond to the time series distances in the data to provide a baseline. 
Ellis et al.~\cite{ellis_novel_2021} can create a time series that can still be considered a close outlier to the distribution. 
However, by using the frequency domain to generate the time series, such a result is expected for most datasets since the correlation between time points is much more important for this approach. 
Despite this, a deep learning model can learn entirely different structures to recognize the class, making such an approach produce a rather misleading time series. 
The other approaches, as seen in\autoref{tab:moa_ts}, perform significantly worse in the Mahalanobis distance for the time series. 
This result makes sense, as the generated time series by the other approaches are quite different from those generated by Ellis et al.~\cite{ellis_novel_2021}.

Our observation was rather correct, as this is not the best approach to compare to the original time series. 
However, we note that the outlier detection works for the activations, which can guide the generation of activation-maximized time series during creation and can be used to extend the \seqdream approach in future works. 
Other possible future extensions for evaluation include using autoencoders for outlier detection or out-of-distribution detection. 
Additionally, other outlier detection mechanisms based on uncertainty quantification toward the activation maximizations can be explored. 
These present possible future research opportunities.

\clearpage

\subsection{Visual out-of-distribution evaluation}

After inspecting the raw numbers and uncovering some interesting findings, we combine the visual observations' results and the numbers from the outlier detection into distribution plots. 
First, we use violin plots to show activations and the results of the generated time series, similar to Ellis et al.~\cite{ellis_novel_2021}. 
Next, we use projections (PCA) of the activations, as seen in~\autoref{fig:general_motivation}.

\begin{adjustbox}{center,caption={Violin distribution plot of the activations for the training data of the model for the class zero on the left and on the right for class one (\underlined{asteroid1}{first class}, \underlined{asteroid2}{second class}). Ellis et al.~\cite{ellis_novel_2021} and \seqdream center both are near the mean of the distribution on both classes, while \seqdream max is above the general activations. Interestingly, with loss shows good results on top of the activations.},label={fig:violin-distribution},nofloat=figure,vspace=\bigskipamount}
\includegraphics[trim=0.7cm 0.2cm 0.7cm 0.2cm,clip,width=0.95\textwidth]{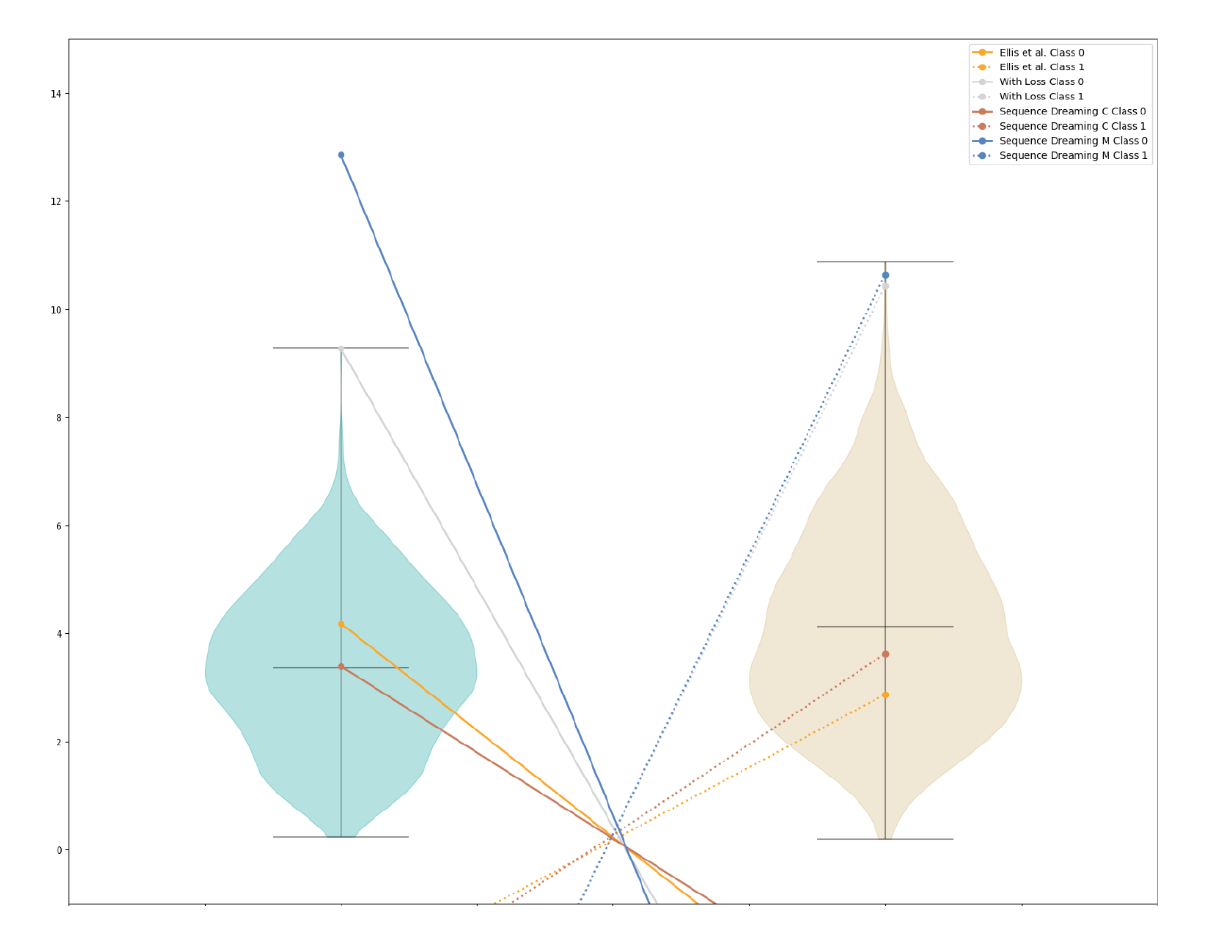}
\end{adjustbox}

The results are unsurprising of~\autoref{fig:violin-distribution} and align with the previous Mahalanobis distance numbers. 
However,~\autoref{fig:violin-distribution} illustrates how closely the activations of the generated time series from Ellis et al.~\cite{ellis_novel_2021} and \seqdream center are aligned. 
Interestingly, the "with loss" approach shows lower activations than the \seqdream max approach. 
We had to exclude the "without loss" data as the activations were too high and would have rendered the distribution plot unreadable without axis scaling. 
Ellis et al.~\cite{ellis_novel_2021} achieves an activation mean similar to that of the training data, suggesting that their method effectively maximizes activation by highlighting patterns of interest to the model. 
However, \seqdream also captures some of these patterns without relying on the frequency domain, producing more abstract, non-periodic results, which indicate some other learned representations by the model unseen to Ellis et al.~\cite{ellis_novel_2021}.
Such a result can indicate that the model learns some values by heart to classify certain samples, which decreases the generalizability of the model.

\begin{adjustbox}{center,caption={First two principal components of a PCA on the activations of the model for the training data for the two classes (\underlined{asteroid1}{first class}, \underlined{asteroid2}{second class}). Afterward, the generated time series are also included. However, we had to exclude Without Loss as these would be outside of the current viewport by a large margin. These would destroy the visualization in general. Our \seqdream approach captures in center and max quite well what we want to have with either class centers or borders. Ellis et al.~\cite{ellis_novel_2021} produce good class centers.},label={fig:pca_with_generated},nofloat=figure,vspace=\bigskipamount}
\includegraphics[trim=0.8cm 0.7cm 0.7cm 0.6cm,clip,width=0.95\textwidth]{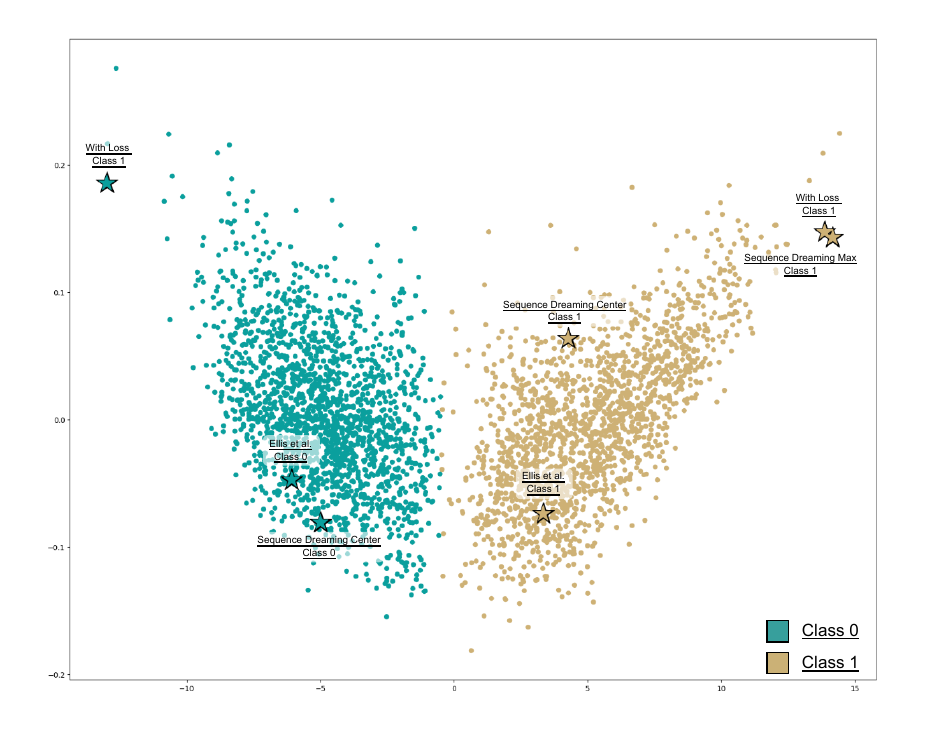}
\end{adjustbox}

As we have seen in previous sections, the activations work quite well in evaluating if the generated time series are positioned as desired. 
We can use the approach shown in~\autoref{fig:general_motivation}. 
By collecting all the activations of a dataset, we can generate a PCA projection using only the first two principal components, as illustrated on the right side of~\autoref{fig:general_motivation}. 
We can then add the generated time series to obtain a figure such as~\autoref{fig:pca_with_generated}. 
Similar to the Mahalanobis distances in~\autoref{tab:moa_act}, we see that Ellis et al.~\cite{ellis_novel_2021} form quite nicely centered activation maximizations, as seen in (A) or (B) of~\autoref{fig:general_approach}. 
\seqdream can achieve similar results with \seqdream center. 
Additionally, \seqdream can achieve maximum and border activations, such as (C) and (D) in~\autoref{fig:general_approach}. 
Thus, with slight changes to the hyperparameters, \seqdream can generate time series for different use cases, revealing some aspects of the inner processes of a deep learning model at the borders, as seen in~\autoref{fig:pca_with_generated}.

\section{Conclusion and Future Work}

In conclusion, we conducted a study on effective regularization for time series classification data without altering the loss function, using only gradient ascent and a modified loss function to achieve good results. 
Based on our observations, we introduced \seqdream, an adaptation of previous methods, incorporating enhanced regularization with increased smoothness. 
The results demonstrate that adding regularization terms to data transformation and the loss function can be effective. 
We learned that activation maximization with regularization terms is highly parameter-sensitive but can produce convincing results comparable to those of frequency domain approaches, which inherently have advantages. 
Our PCA projections of the activations particularly highlight how our approach can generate precise time series corresponding to specific activations with smooth properties.


For future work, several promising directions are poised to refine the \seqdream process for time series data. 
Among these, exploring advanced regularization techniques, such as introducing perturbations directed towards less contributing time points informed by attributions, offers a nuanced method for enhancing model interpretability while minimizing information loss. 
Additionally, the adoption of genetic or evolutionary algorithms, inspired by the works of Nguyen et al.~\cite{nguyen_deep_2015} and Xiao and Kreiman~\cite{xiao_gradient_2019}, presents an intriguing avenue for optimizing activation maximization through a process that mimics natural selection, potentially uncovering novel and highly effective stimuli. 
Moreover, incorporating wavelet transforms, akin to Fourier transformations, could provide a more comprehensive analysis of time series by capturing both the frequency and location in time of significant features, thereby offering a richer representation of the data for time series activation maximization generation. 
Also, incorporating attributions more heavily in the process could lead to more plausible activation maximization time series as these can regularize the generation process.
However, selecting a working attribution technique is not trivial and needs another careful consideration~\cite{schlegel_deep_2023}.


\begin{acknowledgments}
  This work has been partially supported by the Federal Ministry of Education and Research (BMBF) in VIKING (13N16242).
\end{acknowledgments}

\bibliography{main}

\begin{thebibliography}{16}
\expandafter\ifx\csname natexlab\endcsname\relax\def\natexlab#1{#1}\fi
\providecommand{\url}[1]{\texttt{#1}}
\providecommand{\href}[2]{#2}
\providecommand{\path}[1]{#1}
\providecommand{\DOIprefix}{doi:}
\providecommand{\ArXivprefix}{arXiv:}
\providecommand{\URLprefix}{URL: }
\providecommand{\Pubmedprefix}{pmid:}
\providecommand{\doi}[1]{\href{http://dx.doi.org/#1}{\path{#1}}}
\providecommand{\Pubmed}[1]{\href{pmid:#1}{\path{#1}}}
\providecommand{\bibinfo}[2]{#2}
\ifx\xfnm\relax \def\xfnm[#1]{\unskip,\space#1}\fi
\bibitem[{Simonyan et~al.(2014)Simonyan, Vedaldi, and Zisserman}]{simonyan_deep_2014}
\bibinfo{author}{K.~Simonyan}, \bibinfo{author}{A.~Vedaldi}, \bibinfo{author}{A.~Zisserman},
\newblock \bibinfo{title}{Deep inside convolutional networks: visualising image classification models and saliency maps},
\newblock in: \bibinfo{booktitle}{International Conference on Learning Representations (ICLR)}, \bibinfo{year}{2014}.
\bibitem[{Mordvintsev et~al.(2015)Mordvintsev, Olah, and Tyka}]{mordvintsev_deepdream_2015}
\bibinfo{author}{A.~Mordvintsev}, \bibinfo{author}{C.~Olah}, \bibinfo{author}{M.~Tyka},
\newblock \bibinfo{title}{Deepdream-a code example for visualizing neural networks},
\newblock \bibinfo{journal}{Google Research}  (\bibinfo{year}{2015}).
\bibitem[{Olah et~al.(2017)Olah, Mordvintsev, and Schubert}]{olah_feature_2017}
\bibinfo{author}{C.~Olah}, \bibinfo{author}{A.~Mordvintsev}, \bibinfo{author}{L.~Schubert},
\newblock \bibinfo{title}{{Feature Visualization}},
\newblock \bibinfo{journal}{Distill}  (\bibinfo{year}{2017}). \bibinfo{note}{Https://distill.pub/2017/feature-visualization}.
\bibitem[{Theissler et~al.(2022)Theissler, Spinnato, Schlegel, and Guidotti}]{theissler_explainable_2022}
\bibinfo{author}{A.~Theissler}, \bibinfo{author}{F.~Spinnato}, \bibinfo{author}{U.~Schlegel}, \bibinfo{author}{R.~Guidotti},
\newblock \bibinfo{title}{{Explainable AI for Time Series Classification: A review, taxonomy and research directions}},
\newblock \bibinfo{journal}{IEEE Access}  (\bibinfo{year}{2022}).
\bibitem[{Mahendran and Vedaldi(2015)}]{mahendran_understanding_2015}
\bibinfo{author}{A.~Mahendran}, \bibinfo{author}{A.~Vedaldi},
\newblock \bibinfo{title}{Understanding deep image representations by inverting them},
\newblock in: \bibinfo{booktitle}{IEEE Conference on Computer Vision and Pattern Recognition}, \bibinfo{year}{2015}.
\bibitem[{Yosinski et~al.(2015)Yosinski, Clune, Nguyen, Fuchs, and Lipson}]{yosinski_understanding_2015}
\bibinfo{author}{J.~Yosinski}, \bibinfo{author}{J.~Clune}, \bibinfo{author}{A.~Nguyen}, \bibinfo{author}{T.~Fuchs}, \bibinfo{author}{H.~Lipson},
\newblock \bibinfo{title}{Understanding neural networks through deep visualization},
\newblock \bibinfo{journal}{arXiv preprint arXiv:1506.06579}  (\bibinfo{year}{2015}).
\bibitem[{Nguyen et~al.(2015)Nguyen, Yosinski, and Clune}]{nguyen_deep_2015}
\bibinfo{author}{A.~Nguyen}, \bibinfo{author}{J.~Yosinski}, \bibinfo{author}{J.~Clune},
\newblock \bibinfo{title}{Deep neural networks are easily fooled: High confidence predictions for unrecognizable images},
\newblock in: \bibinfo{booktitle}{IEEE Conference on Computer Vision and Pattern Recognition}, \bibinfo{year}{2015}.
\bibitem[{Schlegel et~al.(2023)Schlegel, Oelke, Keim, and El-Assady}]{schlegel_visual_2023}
\bibinfo{author}{U.~Schlegel}, \bibinfo{author}{D.~Oelke}, \bibinfo{author}{D.~A. Keim}, \bibinfo{author}{M.~El-Assady},
\newblock \bibinfo{title}{{Visual Explanations with Attributions and Counterfactuals on Time Series Classification}},
\newblock \bibinfo{journal}{arXiv preprint arXiv:2307.08494}  (\bibinfo{year}{2023}). \href{http://arxiv.org/abs/2307.08494}{{\tt arXiv:2307.08494}}.
\bibitem[{Yoshimura et~al.(2019)Yoshimura, Maekawa, and Hara}]{yoshimura_preliminary_2019}
\bibinfo{author}{N.~Yoshimura}, \bibinfo{author}{T.~Maekawa}, \bibinfo{author}{T.~Hara},
\newblock \bibinfo{title}{Preliminary investigation of visualizing human activity recognition neural network},
\newblock in: \bibinfo{booktitle}{Conference on Mobile Computing and Ubiquitous Network (ICMU)}, \bibinfo{year}{2019}.
\bibitem[{Yoshimura et~al.(2021)Yoshimura, Maekawa, and Hara}]{yoshimura_understanding_2021}
\bibinfo{author}{N.~Yoshimura}, \bibinfo{author}{T.~Maekawa}, \bibinfo{author}{T.~Hara},
\newblock \bibinfo{title}{Toward understanding acceleration-based activity recognition neural networks with activation maximization},
\newblock in: \bibinfo{booktitle}{International Joint Conference on Neural Networks (IJCNN)}, \bibinfo{year}{2021}.
\bibitem[{Ellis et~al.(2021)Ellis, Sendi, Miller, and Calhoun}]{ellis_novel_2021}
\bibinfo{author}{C.~A. Ellis}, \bibinfo{author}{M.~S.~E. Sendi}, \bibinfo{author}{R.~Miller}, \bibinfo{author}{V.~Calhoun},
\newblock \bibinfo{title}{A novel activation maximization-based approach for insight into electrophysiology classifiers},
\newblock in: \bibinfo{booktitle}{IEEE International Conference on Bioinformatics and Biomedicine (BIBM)}, \bibinfo{year}{2021}.
\bibitem[{Kingma and Ba(2014)}]{kingma_adam_2014}
\bibinfo{author}{D.~P. Kingma}, \bibinfo{author}{J.~Ba},
\newblock \bibinfo{title}{Adam: A method for stochastic optimization},
\newblock \bibinfo{journal}{arXiv preprint arXiv:1412.6980}  (\bibinfo{year}{2014}).
\bibitem[{Dau et~al.(2019)Dau, Bagnall, Kamgar, Yeh, Zhu, Gharghabi, Ratanamahatana, and Keogh}]{dau_ucr_2019}
\bibinfo{author}{H.~A. Dau}, \bibinfo{author}{A.~Bagnall}, \bibinfo{author}{K.~Kamgar}, \bibinfo{author}{C.-C.~M. Yeh}, \bibinfo{author}{Y.~Zhu}, \bibinfo{author}{S.~Gharghabi}, \bibinfo{author}{C.~A. Ratanamahatana}, \bibinfo{author}{E.~Keogh},
\newblock \bibinfo{title}{{The UCR time series archive}},
\newblock \bibinfo{journal}{IEEE/CAA Journal of Automatica Sinica} \bibinfo{volume}{6} (\bibinfo{year}{2019}) \bibinfo{pages}{1293--1305}.
\bibitem[{Mahalanobis(1936)}]{mahalanobis_generalized_1936}
\bibinfo{author}{P.~C. Mahalanobis},
\newblock \bibinfo{title}{On the generalized distance in statistics},
\newblock \bibinfo{journal}{Sankhy{\=a}: The Indian Journal of Statistics,}  (\bibinfo{year}{1936}).
\bibitem[{Xiao and Kreiman(2019)}]{xiao_gradient_2019}
\bibinfo{author}{W.~Xiao}, \bibinfo{author}{G.~Kreiman},
\newblock \bibinfo{title}{Gradient-free activation maximization for identifying effective stimuli},
\newblock \bibinfo{journal}{arXiv preprint arXiv:1905.00378}  (\bibinfo{year}{2019}).
\bibitem[{Schlegel and Keim(2023)}]{schlegel_deep_2023}
\bibinfo{author}{U.~Schlegel}, \bibinfo{author}{D.~A. Keim},
\newblock \bibinfo{title}{{A Deep Dive into Perturbations as Evaluation Technique for Time Series XAI}},
\newblock in: \bibinfo{booktitle}{World Conference on Explainable Artificial Intelligence (xAI)}, \bibinfo{year}{2023}.

\end{thebibliography}
\end{document}